\documentclass[10pt,twocolumn,letterpaper]{article}

\usepackage[pagenumbers]{paper} 

\usepackage{graphicx}
\usepackage{amsmath}
\usepackage{amssymb}
\usepackage{booktabs}
\usepackage{multirow}
\usepackage{color}
\usepackage[figuresright]{rotating}

\usepackage[pagebackref,breaklinks,colorlinks]{hyperref}

\usepackage[capitalize]{cleveref}
\crefname{section}{Sec.}{Secs.}
\Crefname{section}{Section}{Sections}
\Crefname{table}{Table}{Tables}
\crefname{table}{Tab.}{Tabs.}

\begin{document}

\title{Multi-View Stereo with Transformer}

\author{{Jie Zhu{$^{1}$}} \quad Bo Peng{$^{1}$} \quad  Wanqing Li{$^{2}$} \quad  Haifeng Shen{$^{3}$} \quad Zhe Zhang{$^{1}$} \quad Jianjun Lei{$^{1}$}\thanks{Corresponding author.}\\
	\normalsize
	$^{1}$\	Tianjin University ~~ $^{2}$\,University of Wollongong ~~ $^{3}$\ AI Labs, Didi Chuxing\\
	\normalsize
	{\tt\small  \{3016207561,bpeng,zz300,jjlei\}@tju.edu.cn\quad wanqing@uow.edu.au\quad
		shenhaifeng@didiglobal.com}
}

\maketitle

\begin{abstract}
   This paper proposes a network, referred to as MVSTR, for Multi-View Stereo (MVS). It is built upon Transformer and is capable of extracting dense features with global context and 3D consistency, which are crucial to achieving reliable matching for MVS. Specifically, to tackle the problem of the limited receptive field of existing CNN-based MVS methods, a global-context Transformer module is first proposed to explore intra-view global context. In addition, to further enable dense features to be 3D-consistent, a 3D-geometry Transformer module is built with a well-designed cross-view attention mechanism to facilitate inter-view information interaction. Experimental results show that the proposed MVSTR achieves the best overall performance on the DTU dataset and strong generalization on the Tanks \& Temples benchmark dataset.
\end{abstract}

\section{Introduction}

As a fundamental problem in computer vision, multi-view stereo (MVS) has received great interest because of its wide range of applications, such as robot navigation, autonomous driving, and augmented reality. Given calibrated multi-view images, MVS aims to recover a 3D model of the observed scene. For decades, MVS has been studied extensively to improve the quality of the recovered 3D models. While traditional methods \cite{tola,camp,furu,gipuma,colmap,acmm} have made great progress, there are still some intractable problems, such as incomplete reconstruction and limited scalability.

\begin{figure}[!t]
	\centering
	\includegraphics[width=1\linewidth]{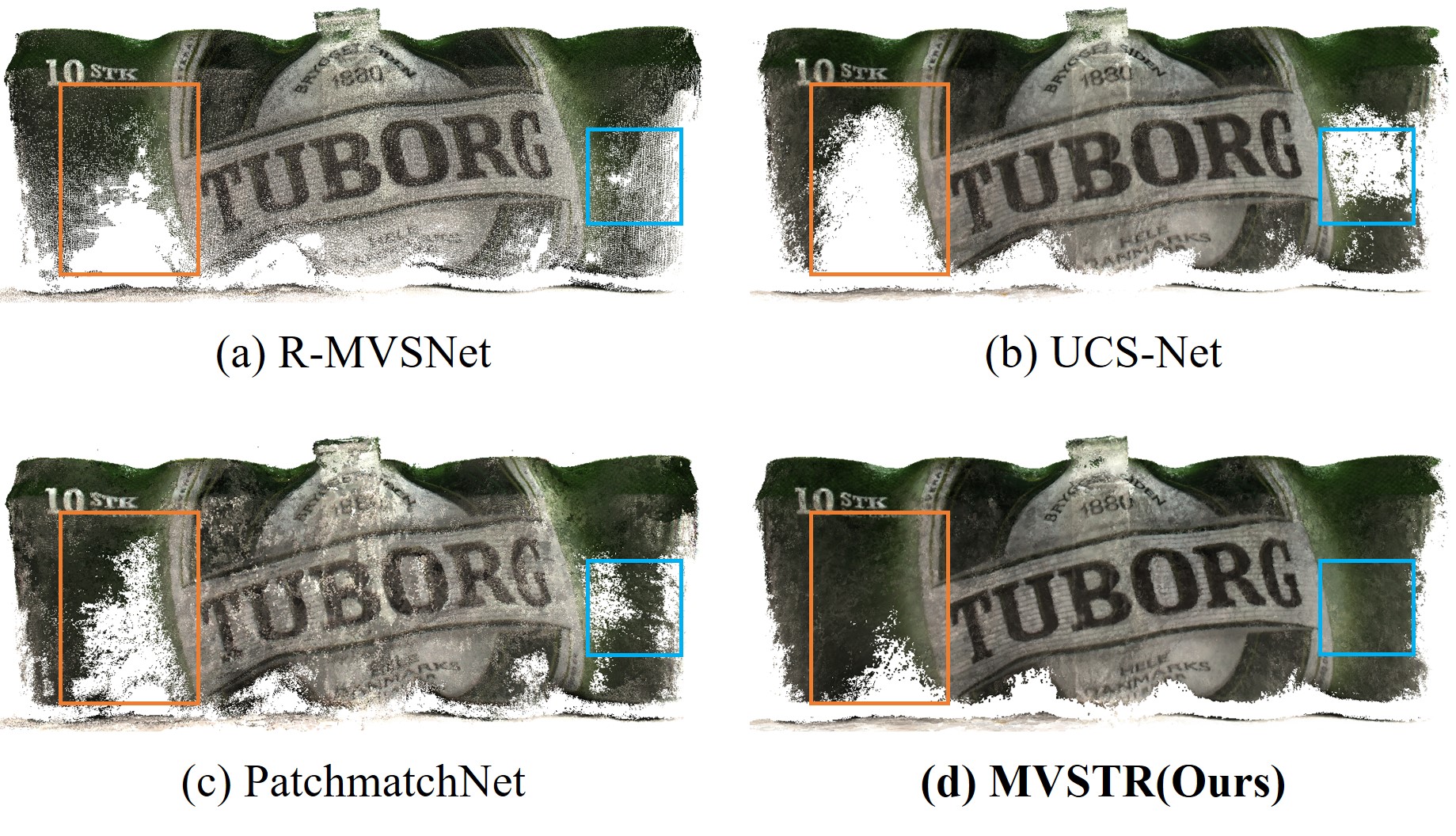}
	\caption{Visual comparison between the CNN-based state-of-the-art MVS methods \cite{rmvsnet,ucsnet,patchmatchnet} and the proposed Transformer-based MVSTR.}
	\label{fig.1}
	\vspace{-0.5cm}
\end{figure}

Recently, learning-based MVS methods \cite{mvsnet,rmvsnet,pmvsnet,mvscrf,pointmvsnet,cider,fastmvsnet,casmvsnet,cvpmvsnet,ucsnet,pvamvsnet,patchmatchnet,aarmvsnet,eppmvsnet} have shown superior performance over traditional counterparts on MVS benchmarks \cite{dtu,t&t}. These learning-based methods make use of convolutional neural networks (CNNs) to infer a depth map for each view, and carry out a separate multi-view depth fusion process to reconstruct 3D point clouds. For depth map inference, they generally first utilize a 2D CNN to extract dense features of each view separately, and then perform robust feature matching to regress the depth map. However, these methods often suffer from matching ambiguities and mismatches in challenging regions, such as texture-less areas or non-Lambertian surfaces. One of the main reasons is that dense features extracted by CNNs with a limited receptive field are difficult to capture global context. The lack of global context usually leads to local ambiguities in untextured or texture-less regions, thus reducing the robustness of matching. Although some recent works \cite{aarmvsnet,mvscrf} try to obtain large context using deformable convolution or multi-scale information aggregation, the solution of mining the global context in each view has not been explored yet for MVS. Besides, in previous methods, the feature of each view is extracted independently from other views. These independently extracted features are hardly optimal for 3D reconstruction. For MVS, due to the widespread existence of non-Lambertian surfaces, the features without being 3D-consistent at the same 3D position may vary considerably across different views, which leads to mismatches. Therefore, exploring a way to access 3D-consistent features is critical for robust and reliable matching in MVS.
 
To cope with the aforementioned problems, a novel Transformer-based network MVSTR is proposed in this paper. Motivated by the long-range modeling capabilities of self-attention in Transformer, a global-context Transformer is designed so that intra-view global context is effectively acquired. In addition, a 3D-geometry Transformer is constructed with cross-view attention. With the well-designed cross-view attention mechanism, the feature of each view will be guided by the information from other views to improve 3D consistency. The proposed MVSTR achieves reliable matching in challenging regions, and thus obtains more accurate and complete reconstruction results than the CNN-based methods, as shown in Figure \ref{fig.1}.

The main contributions of this paper are as follows:

1) A new MVS network built upon Transformer, termed MVSTR, is proposed. To our best knowledge, this is the first Transformer architecture for MVS. 

2) A global-context Transformer is proposed to explore the intra-view global context.

3) To acquire 3D-consistent features, a 3D-geometry Transformer with the well-designed cross-view attention mechanism is proposed to efficiently enable inter-view information interaction for extraction of multi-view features.

4) The proposed method outperforms the state-of-the-art methods on the DTU dataset \cite{dtu} and achieves robust generalization on the Tanks \& Temples benchmark dataset \cite{t&t}.

\section{Related Work}
\label{sec:formatting}

\subsection{Traditional MVS Methods}

In the last decades, traditional MVS methods have been thoroughly researched and made great progress. Furukawa \textit{et al.} \cite{furu} implemented MVS as a procedure of match, expansion, and filter. Campbell \textit{et al.} \cite{camp} designed a discrete label Markov Random Field (MRF) optimization to remove outliers of depth maps. Tola \textit{et al.} \cite{tola} proposed an efficient MVS method, which utilizes DAISY descriptors for matching to reduce computational complexity of large-scale MVS reconstruction. Additionally, a diffusion-like propagation scheme based on the patchmatch stereo algorithm \cite{patchmatchstereo} is employed in \cite{gipuma} to fully utilize parallel acceleration capabilities of GPUs. An MVS system named COLMAP \cite{colmap} is developed to jointly estimate scene depth, surface normal, and pixel-wise view selection. More recently, Xu \textit{et al.} \cite{acmm} improved depth map quality by utilizing structured region information and multi-scale geometric consistency. Although these traditional methods have greatly advanced the development of MVS, their robustness and reliability have been challenged by multiple issues such as low-texture regions, illumination changes, and reflections.

\subsection{Learning-Based MVS Methods}

Inspired by the great success of CNNs in computer vision, many learning-based MVS methods have been proposed recently. Yao \textit{et al.} \cite{mvsnet} proposed MVSNet to infer the depth map for large-scale MVS reconstruction. The MVSNet builds the variance-based cost volume with features extracted by a 2D CNN and performs cost regularization with a 3D CNN to regress the depth map. In order to reduce the huge memory consumption of the 3D CNN, the cost volume is regularized with gated recurrent units (GRUs) \cite{gru} in \cite{rmvsnet}. Besides, Wei \textit{et al.} \cite{aarmvsnet} proposed AA-RMVSNet to further improve the performance of MVS reconstruction by utilizing adaptive aggregation and long short-term memory (LSTM) \cite{lstm}. To develop a computationally efficient network, some works adopt multi-stage strategy to achieve MVS reconstruction. Yu \textit{et al.} \cite{fastmvsnet} designed Fast-MVSNet, which firstly forms a sparse cost volume to infer an initial sparse depth map and then optimizes the sparse depth map gradually. Gu \textit{et al.} \cite{casmvsnet} constructed a cascade cost volume in a coarse-to-fine manner to recover a high-resolution depth map. Cheng \textit{et al.} \cite{ucsnet} built adaptive thin volumes in multiple stages to progressively increase depth resolution and precision.  Moreover, by adopting the pyramid structure, Yang \textit{et al.} \cite{cvpmvsnet} designed CVP-MVSNet to iteratively refine depth maps. Wang \textit{et al.} \cite{patchmatchnet} introduced the traditional patchmatch stereo algorithm \cite{patchmatchstereo} into a learning-based coarse-to-fine framework. To further improve the performance of MVS reconstruction, Ma \textit{et al.} \cite{eppmvsnet} utilized an epipolar-assembling module to regress a coarse depth map and developed an entropy-based strategy for refinement. Though the learning-based methods have achieved great success, they still suffer from match ambiguities and mismatches in challenging regions probably due to insufficient global context and 3D-inconsistent representations. 

\subsection{Transformer in Vision Related Tasks}

\begin{figure*}[!t]
	\centering
	\includegraphics[width=1\linewidth]{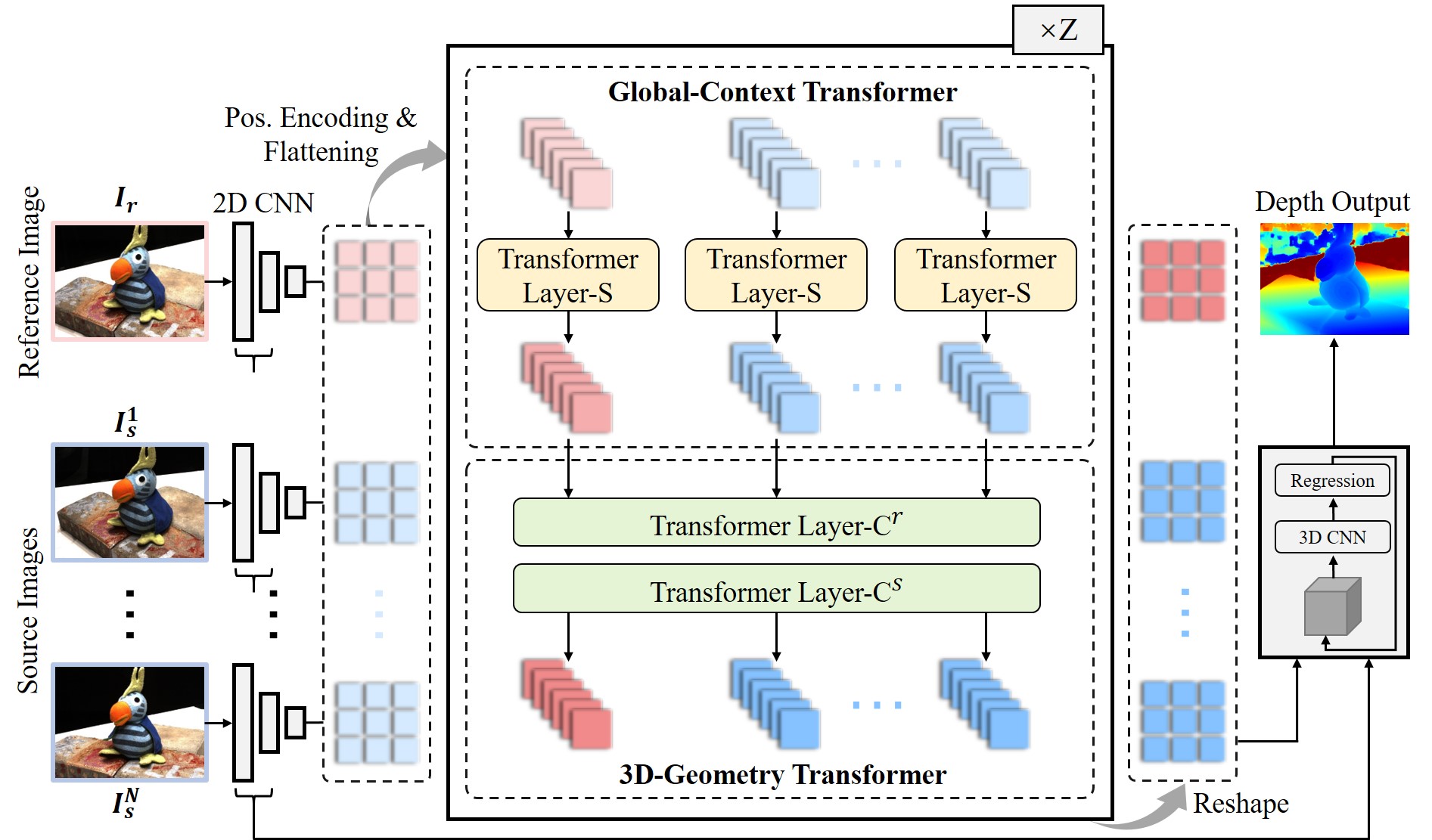}
	\caption{The overall architecture of the proposed MVSTR.}
	\label{fig.2}
\end{figure*}

Thanks to the powerful capabilities of modeling long-range dependencies, Transformer \cite{transformer} and its variants \cite{reformer,lineartransformer,setransformer} have greatly advanced the development of natural language processing. Recently, Transformer has been increasingly explored in various vision tasks. Dosovitskiy \textit{et al.} \cite{vit} proposed a vision Transformer to boost the performance of image classification. Carion \textit{et al.} \cite{detr} presented an end-to-end object detection network based on Transformer. Besides, Transformers are utilized for homography estimation, relative pose estimation, and visual localization in \cite{loftr}. A pre-trained image processing Transformer is introduced for super-resolution, denoising, and deraining in \cite{ipt}. In addition, Li \textit{et al.} \cite{sttr} proposed to take the advantages of both CNN and Transformer for stereo disparity estimation. Liu \textit{et al.} \cite{swin} developed a hierarchical vision Transformer, known as Swin Transformer, for a broad range of vision tasks.

This paper exploits the Transformer architecture for the MVS task. The proposed MVSTR takes full advantages of Transformer to enable features to be extracted under the guidance of global context and 3D geometry, which brings significant improvement on reconstruction results.

\section{Proposed Method}

\subsection{Overall Architecture}
The overall architecture of the proposed MVSTR is illustrated in Figure \ref{fig.2}. Given one reference image $I_{r}$ and $N$ source images $\left\{I_{s}^{i}\right\}_{i=1}^{N}$, local features are first extracted using 2D CNNs and mapped into sequences after positional encoding and flattening. For the feature of each view, the global-context Transformer module is built to explore the intra-view global context. To acquire 3D-consistent dense features, a 3D-geometry Transformer module is built through cross-view attention. With the well-designed cross-view attention mechanism, information interaction across multiple views is efficiently explored. In the proposed MVSTR, the two modules are alternated $Z$ times so that the transformed feature of each view is capable of effectively perceiving both intra-view global context and inter-view 3D geometry. Finally, together with the transformed features and local features, a depth map is generated by a widely-used coarse-to-fine regression strategy \cite{casmvsnet}.

\subsection{Global-Context Transformer}

To acquire dense features with intra-view global information, the proposed global-context Transformer module utilizes multi-head self-attention for long-range dependency learning. Before being fed into the global-context Transformer module, reference feature $F_{r}=\left[F_{r, 1}, F_{r, 2}, \ldots, F_{r, j}\right]$ and source features $\left\{F_{s}^{i}=\left[F_{s, 1}^{i}, F_{s, 2}^{i}, \ldots, F_{s, j}^{i}\right]\right\}_{i=1}^{N}$ extracted from 2D CNNs, are first supplemented with the learnable 2D positional encodings $P=\left[P_{1}, P_{2}, \ldots, P_{j}\right]$ for each pixel, where $j$ denotes the number of pixels of the CNN feature map of each view. It is worth noted that the positional encodings $P$ are the same for all views. Then, the features of the reference view and source views with position information are flattened into sequences $X_{r}$ and $\left\{X_{s}^{i}\right\}_{i=1}^{N}$, which are expressed by:
\begin{equation}
X_{r}=\left[P_{1}+F_{r, 1}, P_{2}+F_{r, 2}, \ldots, P_{j}+F_{r, j}\right]
\label{eq.1}
\end{equation}
\begin{equation}
X_{s}^{i}=\left[P_{1}+F_{s, 1}^{i}, P_{2}+F_{s, 2}^{i}, \ldots, P_{j}+F_{s, j}^{i}\right]
\label{eq.2}
\end{equation}

\begin{figure}[!t]
	\centering
	\includegraphics[width=.96\linewidth]{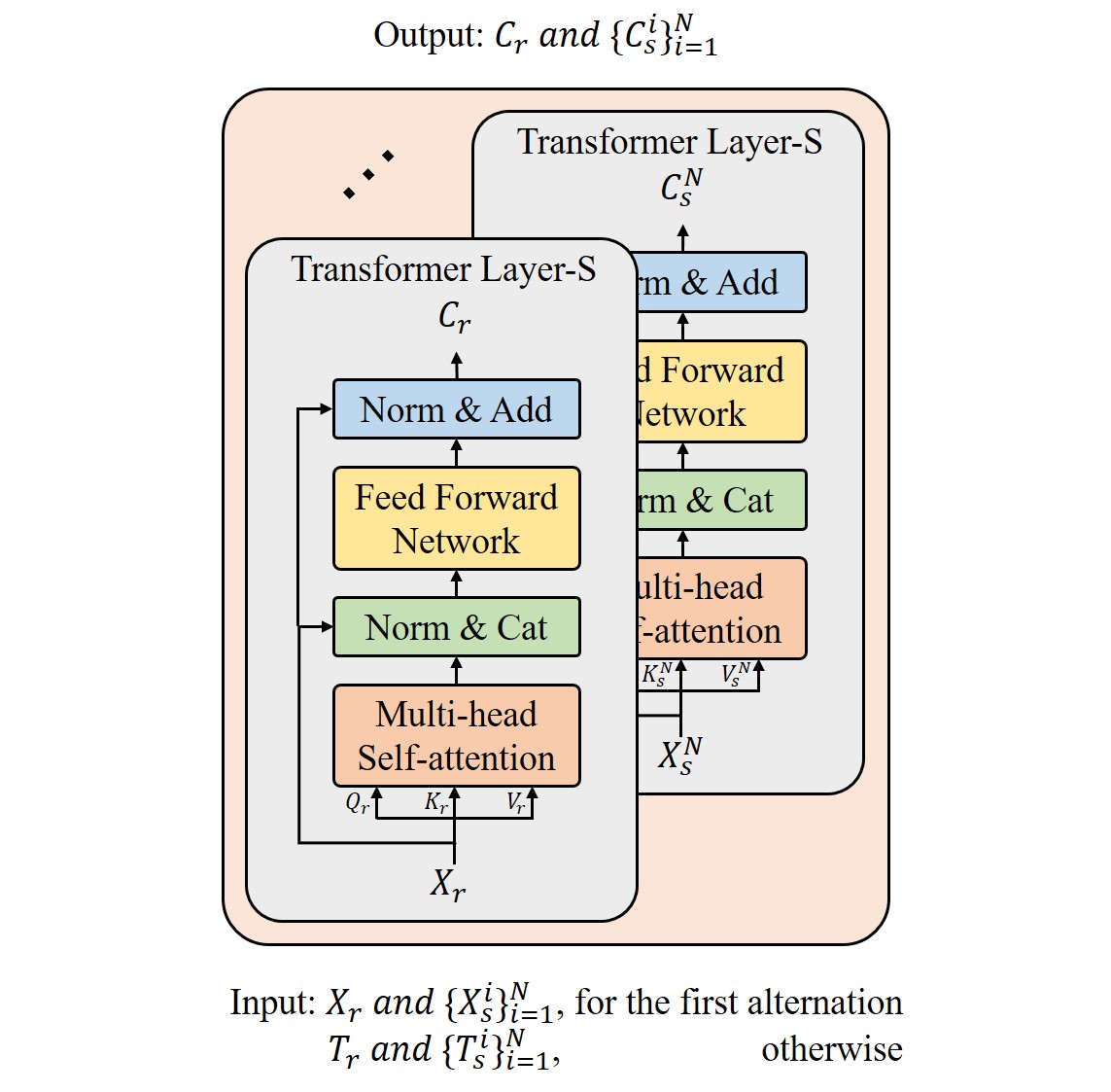}
	\caption{Global-context Transformer module.}
	\label{fig.3}
\end{figure}

The structure of the global-context Transformer module is illustrated as Figure \ref{fig.3}, the core of which is the Transformer Layer-S. For each view, an individual Transformer Layer-S is applied to explore the intra-view global context. Given $X_{r}$ as the input, Transformer Layer-S is expressed as:
\begin{equation}
Q_{r}=K_{r}=V_{r}=X_{r}
\label{eq.3}
\end{equation}
\begin{equation}
X_{r}^{\prime}=\operatorname{Concat}\left(\operatorname{LN}\left(\operatorname{MSA}\left(Q_{r}, K_{r}, V_{r}\right)\right), X_{r}\right)
\label{eq.4}
\end{equation}
\begin{equation}
C_{r}=\operatorname{LN}\left(\operatorname{FFN}\left(X_{r}^{\prime}\right)\right)+X_{r}
\label{eq.5}
\end{equation}
where $\operatorname{Concat}(,)$ denotes the operation of concatenation, $\operatorname{LN}(\cdot)$ denotes the layer normalization. $\operatorname{MSA}\left(Q_{r}, K_{r}, V_{r}\right)$ denotes the multi-head self-attention with query $Q_{r}$, key $K_{r}$, and value $V_{r}$. The multi-head self-attention enables each pixel to establish dependencies with all other pixels within the view. $\operatorname{FFN}(\cdot)$ denotes a fully connected feed-forward network, which is adopted to improve the fitting ability of the model. $C_{r}$ represents the context-aware feature of the reference view. Similarly, given $X_{s}^{i}$ as the input to Transformer Layer-S, the context-aware feature of the corresponding source view $C_{s}^{i}$ is also obtained in the same way.

The global-context Transformer module is capable of exploring the global context within each view. With the context-aware features, the local ambiguities in large untextured or texture-less areas will be reduced.

\subsection{3D-Geometry Transformer}

To acquire dense features with 3D consistency, a 3D-geometry Transformer module is proposed to effectively facilitate the information interaction across multiple views. Employing a cross-view attention mechanism, Transformer Layer-C$^{r}$ is first constructed to obtain 3D-consistent reference feature $T_{r}$ by enabling the reference view to access information in all source views. Then, based on the 3D-consistent reference feature $T_{r}$, Transformer Layer-C$^{s}$ is constructed to acquire features being 3D-consistent with source views by exploiting the information in $T_{r}$.

\begin{figure}[!t]
	\centering
	\includegraphics[width=1\linewidth]{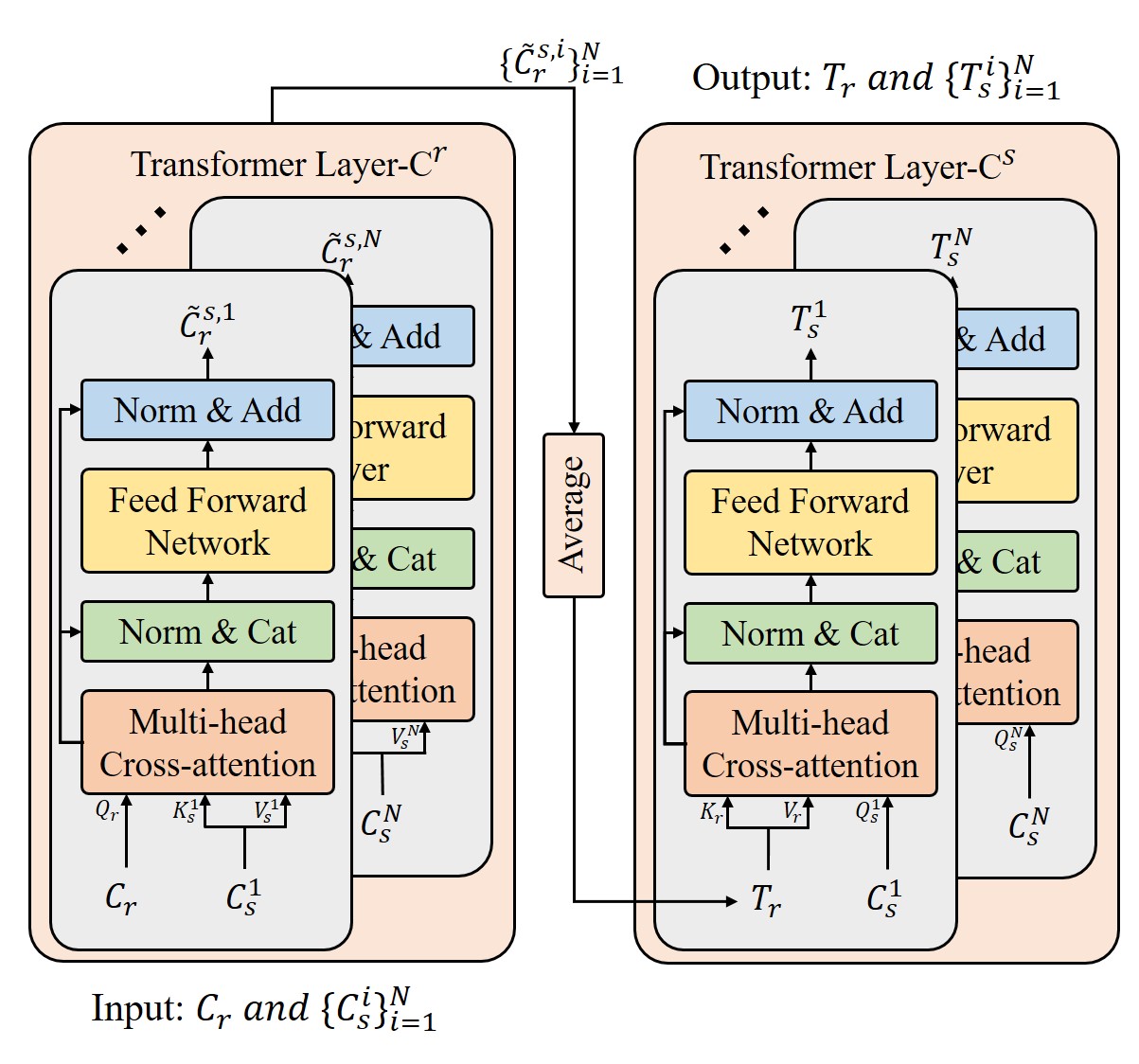}
	\caption{3D-geometry Transformer module.}
	\label{fig.4}
\end{figure}

The details of the 3D-geometry Transformer module with the cross-view attention mechanism are shown in Figure \ref{fig.4}. The context-aware features $C_{r}$ and $\left\{C_{s}^{i}\right\}_{i=1}^{N}$ generated from the global-context Transformer module, are fed into the 3D-geometry Transformer module. To integrate source-view information with the reference view, $N$ cross-view attentions are first employed, each of which is utilized to enhance $C_{r}$ with $C_{s}^{i}$. These $N$ cross-view attentions are denoted as Transformer Layer-C$^{r}$, which is expressed as: 
\begin{equation}
Q_{r}=C_{r}, K_{s}^{i}=V_{s}^{i}=C_{s}^{i}
\label{eq.6}
\end{equation}
\begin{equation}
{C}_{r}^{s, i}=\operatorname{Concat}\left(\operatorname{LN}\left(\operatorname{MCA}\left(Q_{r}, K_{s}^{i}, V_{s}^{i}\right)\right), C_{r}\right)
\label{eq.7}
\end{equation}
\begin{equation}
\widetilde{C}_{r}^{s, i}=\operatorname{LN}\left(\operatorname{FFN}\left(C_{r}^{s, i}\right)\right)+C_{r}
\label{eq.8}
\end{equation}
where $\operatorname{MCA}\left(Q_{r}, K_{s}^{i}, V_{s}^{i}\right))$ denotes the multi-head cross-attention with query $Q_{r}$, key $K_{s}^{i}$, and value $V_{s}^{i}$. $\widetilde{C}_{r}^{s, i}$ denotes the feature of the reference view enhanced by $C_{s}^{i}$.

Based on Transformer Layer-C$^{r}$, $\left\{\widetilde{C}_{r}^{s, i}\right\}_{i=1}^{N}$ are obtained by enhancing $C_{r}$ with $\left\{C_{s}^{i}\right\}_{i=1}^{N}$. Then, an average operation is adopted to fuse the enhanced features $\left\{\widetilde{C}_{r}^{s, i}\right\}_{i=1}^{N}$ to acquire the 3D-consistent feature $T_{r}$ of the reference view, which is formulated as:

\begin{equation}
T_{r}=\frac{1}{N} \sum_{i=1}^{N} \widetilde{C}_{r}^{s, i}
\label{eq.9}
\end{equation}

Subsequently, Transformer Layer-C$^{s}$ is constructed with additional $N$ cross-view attentions, which are utilized to enhance $\left\{C_{s}^{i}\right\}_{i=1}^{N}$ with the 3D-consistent feature $T_{r}$ of the reference view, resulting in 3D-consistent features $\left\{T_{s}^{i}\right\}_{i=1}^{N}$ of source views. The Transformer Layer-C$^{s}$ is formulated as:
\begin{equation}
Q_{s}^{i}=C_{s}^{i}, K_{r}=V_{r}=T_{r}
\label{eq.10}
\end{equation}
\begin{equation}
{C}_{s}^{r, i}=\operatorname{Concat}\left(\operatorname{LN}\left(\operatorname{MCA}\left(Q_{s}^{i}, K_{r}, V_{r}\right)\right), C_{s}^{i}\right)
\label{eq.11}
\end{equation}
\begin{equation}
T_{s}^{i}=\operatorname{LN}\left(\operatorname{FFN}\left({C}_{s}^{r, i}\right)\right)+C_{s}^{i}
\label{eq.12}
\end{equation}
where $\operatorname{MCA}\left(Q_{s}^{i}, K_{r}, V_{r}\right)$ denotes the multi-head cross-attention with query $Q_{s}^{i}$, key $K_{r}$, and value $V_{r}$. $T_{s}^{i}$ represents the 3D-consistent feature of $i$-th source view.

Beneficial from the designed mechanism, 3D-consistent features are obtained for the reference view and source views. With these 3D-consistent features, mismatches in non-Lambertian surfaces are expected to be efficiently alleviated, hence, to improve the 3D reconstruction.

\subsection{Loss Function and Implementation}

{\bf{Loss Function.}} Similar to the existing coarse-to-fine MVS works, smooth $l$1 loss is applied at each scale to supervise depth estimation results of different resolutions \cite{casmvsnet}. The loss function of MVSTR can be formulated as:
\begin{equation}
\text {Loss}=\sum_{m=1}^{M} \alpha_{m} \cdot L_{m}
\label{eq.13}
\end{equation}
where $M$ refers to total number of scales in the proposed network and is set to 3. $L_{m}$ and $\alpha_{m}$ refers to the loss and its corresponding loss weight at scale $m$, respectively. In particular,  $m=1$ represents the coarsest scale while $m=3$ represents the finest scale. With $m$ increasing from 1 to 3, $\alpha_{m}$ is set to 0.5, 1.0, and 2.0, respectively.

\begin{figure*}[!t]
	\centering
	\includegraphics[width=1\linewidth]{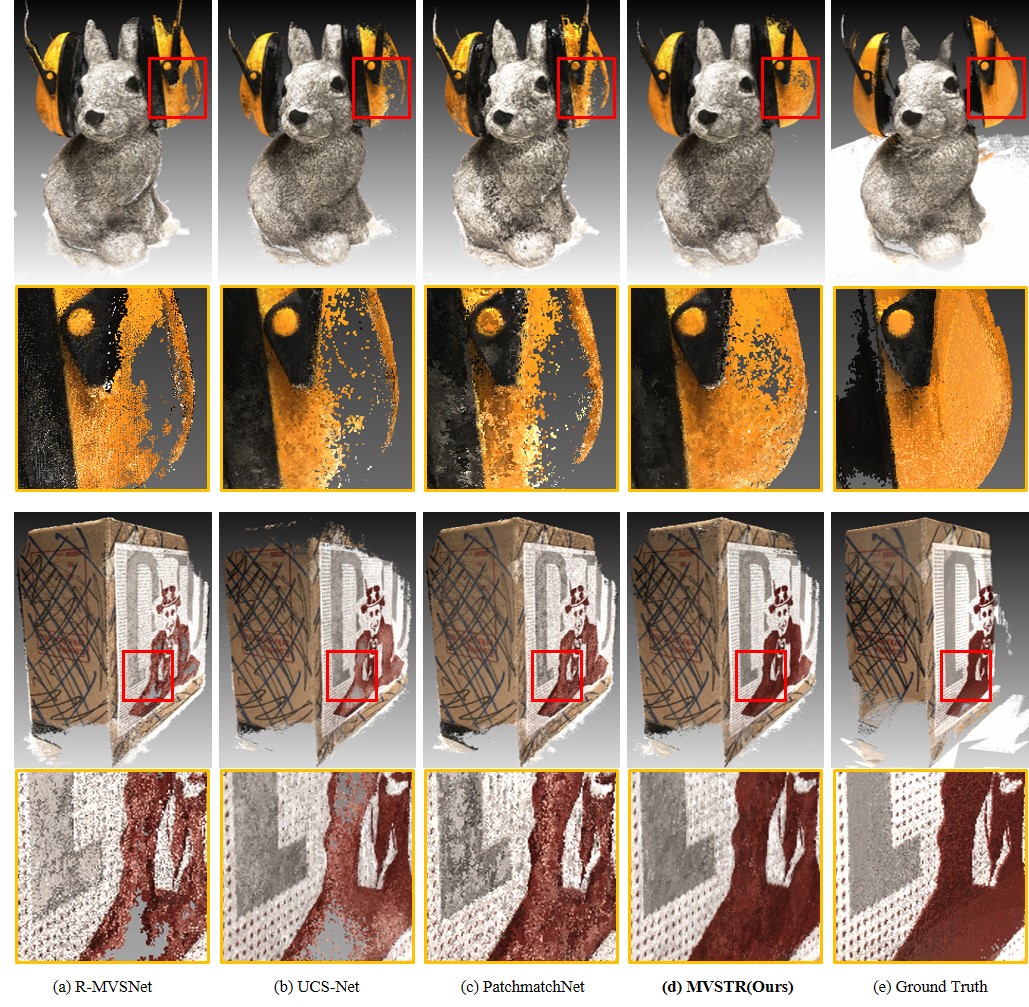}
	\caption{Visual comparison with state-of-the-art methods \cite{rmvsnet,ucsnet,patchmatchnet} of scan33 and scan13 on the DTU evaluation dataset. The first and third rows show the point clouds reconstructed by the corresponding methods while the second and fourth rows show those zoomed-in local areas marked with red rectangle boxes.}
	\label{fig.5}
	\vspace{0.2cm}
\end{figure*}

{\bf{Implementation.}} The 2D CNN which is adopted for extracting feature from individual view, is an eight-layer architecture similar to the one used in MVSNet \cite{mvsnet}. In particular, for the sake of computing efficiency, the batch normalization layer and the ReLU activation are replaced with a unified in-place activated batch normalization layer \cite{inplace-abn}, which brings nearly 40\% memory savings. The output for each view is a 32-channel feature map downsized by four compared with the input image. Note that the weights of the 2D CNNs are shared among multiple views. In addition, the global-context Transformer module and 3D-geometry Transformer module are alternatively stacked $Z=4$ times. For the global-context Transformer module, the weights of Transformer Layer-S are shared among multiple views. Similarly, for both Transformer Layer-C$^{r}$ and Transformer Layer-C$^{s}$, the weights of cross-view attentions are shared among multiple views. For all Transformer layers, a linearized multi-head attention \cite{lineartransformer} is adopted, in which the number of heads is set to 4. Besides, the coarse-to-fine depth regression consists of cost volume pyramid construction and 3D CNN regularization of 3 scales. For cost volume pyramid construction, the transformed features are first utilized to construct cost volume at the coarsest scale via differentiable homography warping \cite{mvsnet} and average group-wise correlation \cite{cider} in which the number of groups is set to 8. At larger scales, the transformed features are first upsampled by bilinear interpolation and fused with the features of the corresponding scale in the 2D CNNs by a convolutional layer with a 1×1 filter. Then, finer cost volumes are constructed with the fused features in the same manner as the coarsest scale. The number of depth hypotheses and the corresponding depth intervals are set to be the same as those in CasMVSNet \cite{casmvsnet}. For 3D CNN cost regularization, 3D U-Nets \cite{unet} without shared weights are applied at 3 scales. Similar to the 2D CNNs, the batch normalization layer and the ReLU activation in 3D U-Nets are replaced with the in-place activated batch normalization layer. Finally, the soft argmin operation \cite{softargmin} is used to regress depth maps at different scales.
\begin{table}[t]
	\centering
	\resizebox{1\linewidth}{!}{
		\begin{tabular}{ccccc}
			\toprule
			\multicolumn{2}{c}{Method} & Acc.(mm) & Comp.(mm) & \textbf{Overall(mm)} \\
			\midrule
			\multirow{5}[2]{*}{\begin{sideways}Traditional\end{sideways}} & Furu \cite{furu} & 0.613  & 0.941  & 0.777  \\
			& Tola \cite{tola} & 0.342  & 1.190  & 0.766  \\
			& Camp \cite{camp} & 0.835  & 0.554  & 0.695  \\
			& Gipuma \cite{gipuma} & \textbf{0.283}  & 0.873  & 0.578  \\
			& Colmap \cite{colmap} & 0.400  & 0.644  & 0.532  \\
			\midrule
			\multirow{13}[2]{*}{\begin{sideways}Learning-based\end{sideways}} & SurfaceNet \cite{surfacenet} & 0.450  & 1.040  & 0.745  \\
			& MVSNet \cite{mvsnet} & 0.396  & 0.527  & 0.462  \\
			& R-MVSNet \cite{rmvsnet} & 0.383  & 0.452  & 0.417  \\
			& P-MVSNet \cite{pmvsnet} & 0.406  & 0.434  & 0.420  \\
			& Point-MVSNet \cite{pointmvsnet} & 0.342  & 0.411  & 0.376  \\
			& CIDER \cite{cider} & 0.417  & 0.437  & 0.427  \\
			& Fast-MVSNet \cite{fastmvsnet} & 0.336  & 0.403  & 0.370  \\
			& CasMVSNet \cite{casmvsnet} & 0.325  & 0.385  & 0.355  \\
			& UCS-Net \cite{ucsnet} & 0.338  & 0.349  & 0.344  \\
			& CVP-MVSNet \cite{cvpmvsnet} & 0.296  & 0.406  & 0.351  \\
			& PVA-MVSNet \cite{pvamvsnet} & 0.379  & 0.336  & 0.357  \\
			& PatchmatchNet \cite{patchmatchnet} & 0.427  & \textbf{0.277}  & 0.352  \\
			& AA-RMVSNet \cite{aarmvsnet} & 0.376  & 0.339  & 0.357  \\
			& EPP-MVSNet \cite{eppmvsnet} & 0.413  & 0.296  & 0.355  \\
			\midrule
			\multicolumn{2}{c}{\textbf{MVSTR(Ours)}} & 0.356  & 0.295  & \textbf{0.326}  \\
			\bottomrule
		\end{tabular}%
	}
	\caption{Quantitative results of different methods on the DTU evaluation set (lower is better).}
	\label{tab.1}%
	\vspace{0.2cm}
\end{table}%
The proposed method is implemented using PyTorch on a machine with a GPU of NVIDIA GeForce GTX 1080Ti and a CPU of Intel Core i9-9900K processor @3.60 GHz. During training, the number of source images $N$ is set to 2, and the resolution of input images is set to 640 × 512. The network is optimized by Adam with $\beta_{1}=0.9$ and $\beta_{2}=0.999$ for 16 epochs. The initial learning rate is set to $1 \times 10^{-3}$ and reduced by half at the 10th, 12th, and 14th epochs.

\section{Experimental Results}

\subsection{Experimental Settings}
\begin{table*}[htbp]
	\centering
	\resizebox{1\linewidth}{!}{
		\begin{tabular}{c|cccccccc|c|cccccc|c}
			\hline
			\multirow{2}{*}{F-score} & \multicolumn{9}{c|}{Intermediate Group}  & \multicolumn{7}{c}{Advanced Group} \\
			\cline{2-17}          
			& Fam.  & Franc. & Horse & Light. & M60   & Pan.  & Play. & Train & \textbf{Mean} & Audi. & Ballr. & Courtr. & Museum & Palace & Temple & \textbf{Mean}\\
			\hline
			MVSNet \cite{mvsnet} & 55.99  & 28.55  & 25.07  & 50.79  & 53.96  & 50.86  & 47.90  & 34.69  & 43.48  &  -    &  -    &  -    &  -    &  -    &  -    &  - \\
			R-MVSNet \cite{rmvsnet} & 69.96  & 46.65  & 32.59  & 42.95  & 51.88  & 48.80  & 52.00  & 42.38  & 48.40  & 12.55  & 29.09  & 25.06  & 38.68  & 19.14  & 24.96  & 24.91  \\
			P-MVSNet \cite{pmvsnet} & 70.04  & 44.64  & 40.22  & 65.20  & 55.08  & 55.17  & 60.37  & 54.29  & 55.62  &  -    &  -    &  -    &  -    &  -    &  -    &  - \\
			Point-MVSNet \cite{pointmvsnet} & 61.79  & 41.15  & 34.20  & 50.79  & 51.97  & 50.85  & 52.38  & 43.06  & 48.27  &  -    &  -    &  -    &  -    &  -    &  -    &  - \\
			CIDER \cite{cider} & 56.79  & 32.39  & 29.89  & 54.67  & 53.46  & 53.51  & 50.48  & 42.85  & 46.76  & 12.77  & 24.94  & 25.01  & 33.64  & 19.18  & 23.15  & 23.12  \\
			Fast-MVSNet \cite{fastmvsnet} & 65.18  & 39.59  & 34.98  & 47.81  & 49.16  & 46.20  & 53.27  & 42.91  & 47.39  &  -    &  -    &  -    &  -    &  -    &  -    &  - \\
			CasMVSNet \cite{casmvsnet} & 76.37  & 58.45  & 46.26  & 55.81  & 56.11  & 54.06  & 58.18  & 49.51  & 56.84  & 19.81  & 38.46  & 29.10  & 43.87  & 27.36  & 28.11  & 31.12  \\
			UCS-Net \cite{ucsnet} & 76.09  & 53.16  & 43.03  & 54.00  & 55.60  & 51.49  & 57.38  & 47.89  & 54.83  &  -    &  -    &  -    &  -    &  -    &  -    &  - \\
			CVP-MVSNet \cite{cvpmvsnet} & 76.50  & 47.74  & 36.34  & 55.12  & 57.28  & 54.28  & 57.43  & 47.54  & 54.03  &  -    &  -    &  -    &  -    &  -    &  -    &  - \\
			PVA-MVSNet \cite{pvamvsnet} & 69.36  & 46.80  & 46.01  & 55.74  & 57.23  & 54.75  & 56.70  & 49.06  & 54.46  &  -    &  -    &  -    &  -    &  -    &  -    &  - \\
			PatchmatchNet \cite{patchmatchnet} & 66.99  & 52.64  & 43.24  & 54.87  & 52.87  & 49.54  & 54.21  & 50.81  & 53.15  & 23.69  & 37.73  & 30.04  & 41.80  & 28.31  & 32.29  & 32.31  \\
			\textbf{MVSTR(Ours)} & 76.92  & 59.82  & 50.16  & 56.73  & 56.53  & 51.22  & 56.58  & 47.48  & \textbf{56.93}  & 22.83  & 39.04  & 33.87  & 45.46  & 27.95  & 27.97  & \textbf{32.85} \\
			\hline
		\end{tabular}%
	}
	\caption{Quantitative results of different methods on the Tanks \& Temples benchmark dataset (higher is better).}
	\label{tab.2}%
	\vspace{0.2cm}
\end{table*}%

\begin{figure*}[!]
	\centering
	\includegraphics[width=1\linewidth]{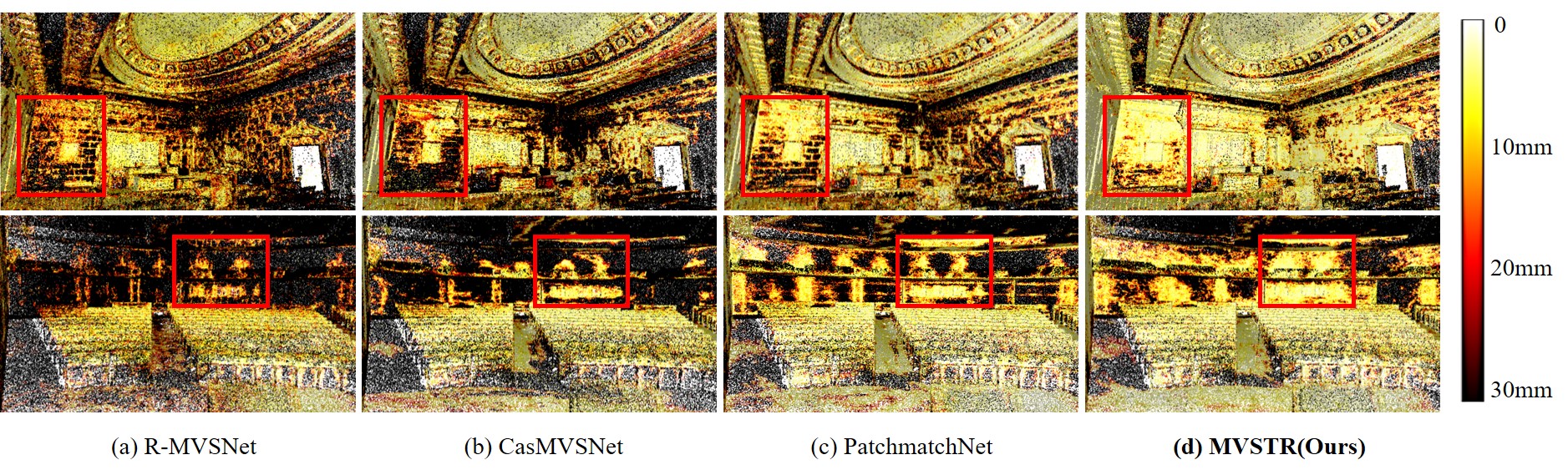}
	\caption{Visual comparison with state-of-the-art methods \cite{rmvsnet,casmvsnet,patchmatchnet} of Courtroom and Auditorium on the Tanks \& Temples benchmark dataset. The top row is the error visualization of Courtroom and the bottom row is the error visualization of Auditorium. The darker in the map the bigger the error in the point cloud.}
	\label{fig.6}
	\vspace{0.2cm}
\end{figure*}

The MVS datasets including DTU \cite{dtu} and Tanks \& Temples \cite{t&t} are used to evaluate the proposed MVSTR. DTU is an indoor dataset, which consists of 124 different scenes. Each scene is covered by 49 or 64 views under 7 different lighting conditions. Tanks \& Temples is a benchmark dataset with both indoor and outdoor scenes. It contains two groups called the intermediate and the advanced ones. The two groups of Tanks \& Temples are used for generalization verification of the proposed method.

During testing on the DTU evaluation set, the number of source images $N$ is set to 4, and the resolution of input images is set to $1152\times864$. When verifying generalization on the Tanks \& Temples benchmark dataset, the number of source images $N$ is set to 6, the input resolution is set to $1920\times1056$ for images with original resolution of $1920\times1080$, and $2048\times1056$ for images with original resolution of $2048\times1080$, which is the same as \cite{casmvsnet}.

Similar to the previous MVS methods \cite{mvsnet, casmvsnet, patchmatchnet}, post-processing steps including photometric depth map filtering, geometric depth map filtering, and depth fusion are used to generate 3D point clouds. The standard evaluation protocol \cite{dtu} is adopted to evaluate the performance of reconstructing 3D point clouds. In particular, the accuracy and the completeness of the reconstructed point clouds are calculated using the official MATLAB code provided by DTU. Further, the average of the accuracy and the completeness is expressed as the overall score. For the Tanks \& Temples benchmark dataset, reconstructed point clouds are uploaded online to calculate F-score. 

\subsection{Evaluation on the DTU Dataset}

To demonstrate the effectiveness of the proposed method, quantitative and qualitative experiments on the DTU dataset are conducted. Table \ref{tab.1} reports the results and comparison with different methods. It can be seen from the table that the proposed MVSTR achieves the best overall performance with competitive accuracy and completeness. In particular, compared with the multi-stage CNN-based methods (\textit{i.e.}, FastMVSNet \cite{fastmvsnet}, CasMVSNet \cite{casmvsnet}, UCS-Net \cite{ucsnet}, CVP-MVSNet \cite{cvpmvsnet}, PatchmatchNet \cite{patchmatchnet}, and EPP-MVSNet \cite{eppmvsnet}), the proposed MVSTR achieves obvious improvement in terms of the overall score. In addition, the proposed MVSTR outperforms the methods with RNN-based cost regularizations (\textit{i.e.}, R-MVSNet \cite{rmvsnet} and AA-RMVSNet \cite{aarmvsnet}) in terms of accuracy, completeness, and the overall score.

Visual comparison of the results achieved by different methods are shown in Figure \ref{fig.5}. It can be seen that the proposed MVSTR reconstructs more complete point clouds with well-preserved structure. Specially, compared with PatchmatchNet \cite{patchmatchnet}, the proposed MVSTR obtains denser reconstruction results with fewer outliers. It is mainly because that the proposed MVSTR is capable of acquiring context-aware and 3D-consistent features, which help reduce matching ambiguities and mismatches in challenging regions to further improve the reconstruction quality.

\subsection{Evaluation on the Tanks \& Temples Benchmark}

\begin{table}[htbp]
	\centering
	\resizebox{1\linewidth}{!}{
		\begin{tabular}{cccc}
			\toprule
			Methods & Acc.(mm) & Comp.(mm) & Overall(mm) \\
			\midrule
			w/o GCT  & 0.369  & 0.297  & 0.333  \\
			w/o 3GT  & 0.363  & 0.298  & 0.331  \\
			MVSTR & \textbf{0.356}  & \textbf{0.295}  & \textbf{0.326}  \\
			\bottomrule
		\end{tabular}%
	}
	\caption{Ablation study of the two Transformer modules (lower is better).}
	\label{tab.3}%
\end{table}%

\begin{table}[htbp]
	\centering
	\resizebox{1\linewidth}{!}{
		\begin{tabular}{cccc}
			\toprule
			Number of Transformers & Acc.(mm) & Comp.(mm) & Overall(mm) \\
			\midrule
			0     & 0.374  & 0.305  & 0.340  \\
			2     & 0.368  & \textbf{0.289}  & 0.329  \\
			4     & \textbf{0.356}  & 0.295  & \textbf{0.326}  \\
			6     & 0.359  & 0.300  & 0.330  \\
			\bottomrule
		\end{tabular}%
	}
	\caption{Ablation study of the number of Transformers (lower is better).}
	\label{tab.4}%
\end{table}%

\begin{table}[htbp]
	\centering
	\resizebox{1\linewidth}{!}{
		\begin{tabular}{cccc}
			\toprule
			Method & GPU Memory(MB) & Run-time(s) & Overall(mm) \\
			\midrule
			MVSNet \cite{mvsnet} & 10632 & 1.435 & 0.551 \\
			CasMVSNet \cite{casmvsnet} & 5667  & 0.459 & 0.355 \\
			PatchmatchNet \cite{patchmatchnet} & \textbf{2323} & \textbf{0.417} & 0.374 \\
			\textbf{MVSTR(Ours)} & 3879  & 0.818 & \textbf{0.326} \\
			\bottomrule
		\end{tabular}%
	}
	\caption{Comparison of GPU memory, run-time and overall performance between the proposed method and other state-of-the-art learning-based methods (lower is better).}
	\label{tab.5}%
\end{table}%

In order to verify the generalization of the proposed method, the model trained on the DTU training dataset without any fine-tuning is utilized to test on the Tanks \& Temples benchmark dataset. Table \ref{tab.2} shows the quantitative results of the reconstructed point clouds obtained by the proposed MVSTR and other state-of-the-art methods on both the intermediate and the advanced groups. For a fair comparison, the methods that are only trained on the DTU training dataset are listed in the table. It can be seen from the table that the proposed MVSTR achieves state-of-the-art performance in the intermediate group. For the advanced group, which is more difficult due to complex geometric layouts and camera trajectories, the proposed MVSTR obtains the highest mean F-score among all the methods, which demonstrates the strong generalization of the proposed MVSTR. 

The error maps of the point clouds reconstructed by different methods are visualized in Figure \ref{fig.6}. In contrast to these state-of-the-art methods, the proposed MVSTR reconstructs more accurate and complete point clouds (see the red box in Figure \ref{fig.6} (d)). This mainly benefits from the features with global context and 3D consistency extracted by the proposed MVSTR, which are more suitable to achieve reliable matching on such a difficult benchmark dataset.

\subsection{Ablation Study}

{\bf{Transformer Modules.}} To evaluate the contribution of the proposed global-context Transformer module and 3D-geometry Transformer module, ablation studies are conducted. Experimental results are shown in Table \ref{tab.3}, where ``w/o GCT" refers to the variant removing the global-context Transformer module from MVSTR, and ``w/o 3GT" refers to the variant removing the 3D-geometry Transformer module from MVSTR. It can be seen from the table that both the modules contribute to the improvement of the accuracy and completeness of the reconstructed point clouds. Benefiting from combining the two modules, the proposed MVSTR achieves state-of-the-art performance on the DTU evaluation set.

{\bf{The Number of Transformers.}} To investigate the impact of the number $Z$ of Transformers, Transformers (\textit{i.e.}, the global-context Transformer module and the 3D-geometry Transformer module) are repeated different times. Experimental results on the DTU evaluation set are shown in Table \ref{tab.4}. It can be seen from the table that the overall performance improves with the increase of the number until it reaches 4. However, when the number increases to 6, the performance declines probably because of a larger number of parameters which are more difficult for training process.

\subsection{Complexity Analysis}
MVSNet \cite{mvsnet}, CasMVSNet \cite{casmvsnet} and PatchmatchNet \cite{patchmatchnet} are compared with the proposed MVSTR in terms of memory and run-time performance. For a fair comparison, a fixed input size of $1152\times864$ is used to evaluate the computational cost on a single GPU of NVIDIA GeForce GTX 1080Ti. As shown in Table \ref{tab.5}, the proposed MVSTR achieves 63.5\% memory savings and 43.0\% run-time reduction compared with MVSNet \cite{mvsnet}. Though there is no computational savings compared to PatchmatchNet \cite{patchmatchnet}, the proposed MVSTR achieves obvious improvement on the quality of reconstructed point clouds with acceptable extra memory and run-time.  

\section{Conclusion}
In this paper, a new MVS network built upon Transformer, termed MVSTR, is proposed. Compared with the existing CNN-based MVS networks, the proposed MVSTR enables features to be extracted under the guidance of global context and 3D geometry via the proposed global-context Transformer and 3D-geometry Transformer modules, respectively. Benificial from the two modules, dense features acquired by the proposed MVSTR are more suitable for reliable matching, especially in texture-less regions and non-Lambertian surfaces. Extensive experimental results  demonstrate that the proposed MVSTR outperforms other CNN-based state-of-the-art methods. For real-time applications, the proposed MVSTR still needs to be further improved on efficiency due to the limitation of time complexity.

{\small
\bibliographystyle{ieee_fullname}
\bibliography{egbib}
}

\end{document}